%% file: SterckxKBS_2016.tex
\documentclass[review]{elsarticle}

\usepackage{tikz,pgfplots}
\usepackage[inline,shortlabels]{enumitem}
\usepackage{lineno,hyperref}
\usepackage[english]{babel}
\usepackage[utf8x]{inputenc}
\usepackage{amsmath,amsthm}
\usepackage{amssymb}% http://ctan.org/pkg/amssymb
\usepackage{pifont}% http://ctan.org/pkg/pifont
\usepackage{geometry}
\usepackage{float,lscape}
\usepackage{rotating}
\usepackage{tikz-dependency}
\theoremstyle{plain}
\newtheorem{thm}{Theorem} % reset theorem numbering for each chapter
\theoremstyle{definition}
\newtheorem{resq}[thm]{RQ} % definition numbers are dependent on theorem numbers
\modulolinenumbers[5]
\usetikzlibrary{arrows,shapes} 
\usepackage{times}
\usepackage{qtree}
\usepackage{framed}
\usepackage{verbatim}
\usepackage{pbox}
\usepackage{multirow,siunitx}
\usepackage{soul} % for \st (strikethrough)
\usepackage{booktabs} % for \toprule \midrule \bottomrule

\usepackage{makecell}
\usepackage{subfig}

\newcommand{\inlinelist}[2]{\begin{enumerate*}[label=#1] #2 \end{enumerate*}}

\usepackage[colorinlistoftodos]{todonotes}

\newcommand{\revision}[1]{\textcolor{black}{#1}}
\newcommand{\delete}[1]{\textcolor{red}{\st{#1}}}

\usepackage{color}

\def \eg {e.g., }
\def \ie {i.e., }

\def \etal {et al.\ }
\def \Fone {{$\text{F}_1$ }}

\journal{Knowledge Based Systems}

\begin{document}
\begin{frontmatter}
\title{Knowledge Base Population using Semantic Label Propagation}
\author{Lucas Sterckx}
\cortext[mycorrespondingauthor]{Corresponding author}
\ead{lucas.sterckx@intec.ugent.be}
\author{Thomas Demeester}
\author{Johannes Deleu}
\author{Chris Develder}
%\address{Ghent University -- iMinds \\Gaston Crommenlaan 8, BE-9000 Ghent, Belgium}
\address{Ghent University -- iMinds \\Technologiepark Zwijnaarde 15, BE-9052 Ghent, Belgium}

\begin{abstract}
A crucial aspect of a knowledge base population system that extracts new facts from text corpora, is the generation of training data for its relation extractors. In this paper, we present a method that maximizes the effectiveness of newly trained relation extractors at a minimal annotation cost.
Manual labeling can be significantly reduced by Distant Supervision (DS), which is a method to construct training data automatically by aligning a large text corpus with an existing knowledge base of known facts. For example, all sentences mentioning both `Barack Obama' and `US' may serve as positive training instances for the relation {\it born\_in(subject,object)}.
However, distant supervision typically results in a highly noisy training set: many training sentences containing the known entity pairs do not really express the intended relation. 
\revision{We explore the idea of combining DS with (partial) human supervision to eliminate that noise. This idea is not novel per se, but our key contributions are:
\inlinelist{(\roman*)}{
\item a novel method of filtering the DS training set based on labeling Shortest Dependency Paths, (SDPs), and
\item the Semantic Label Propagation (SLP) model.
}}
We propose to combine DS with minimal manual human supervision \revision{by annotating features (in particular SDPs) rather than (potential) relation instances. Such so-called feature labeling is adopted} to eliminate noise from the large and noisy initial training set, resulting in a significant increase of precision \revision{(at the expense of recall)}. 
%First we demonstrate the effectiveness of feature labeling after initial training using Distant Supervision. 
We further improve on this approach by introducing the Semantic Label Propagation (SLP) method, which uses the similarity between low-dimensional representations of candidate training instances, to extend the (filtered) training set in order to increase recall 
while maintaining high precision.

Our proposed strategy for generating training data is studied and evaluated on an established test collection designed for knowledge base population (KBP) tasks from the TAC KBP English slot filling task.
The experimental results show that the SLP strategy leads to substantial performance gains when compared to existing approaches, while requiring an almost negligible human annotation effort. 
\end{abstract}

\end{frontmatter}
% \linenumbers

\section{Introduction}
In recent years we have seen significant advances in the creation of large-scale Knowledge Bases (KBs), databases containing millions of facts about persons, organizations, events, products, etc. Examples include Wikipedia-based KBs (e.g., YAGO~\cite{suchanek2007yago}, DBpedia~\cite{bizer2009dbpedia}, and Freebase~\cite{bollacker2008freebase}), KBs generated from Web documents (e.g., NELL~\cite{mitchell2015never}, PROSPERA\cite{nakashole2011scalable}), or open information extraction approaches (e.g., TextRunner~\cite{yates2007textrunner}, PRISMATIC~\cite{fan2010prismatic}). 
\revision{
Other Knowledge Bases like Conceptnet~\cite{liu2004conceptnet} or SenticNet~\cite{cambria2010senticnet} collect conceptual information conveyed by natural language and store them in a form which makes them easily accessible to machines.}
Besides the academic projects, several commercial projects were initiated by major corporations like Microsoft (Satori\footnote{\url{https://blogs.bing.com/search/2013/03/21/understand-your-world-with-bing}}), Google (Knowledge Graph~\cite{dong2014knowledge}), Facebook\footnote[2]{\url{http://www.insidefacebook.com/2013/01/14/}}, Walmart~\cite{Deshpande2011} and others. This is driven by a wide variety of applications for which KBs are increasingly found to be essential, \eg digital assistants, or for enhancing search engine results with semantic search information.

\par Because KBs are often manually constructed, they tend to be incomplete. For example, 78.5\% of \textit{persons} in Freebase have no known \textit{nationality} \cite{min2013distant}.
To complete a KB, we need a Knowledge Base Population (KBP) system that extracts information from various sources, of which a large fraction comprises unstructured written text items~\cite{dong2014knowledge}. A vital component of a KBP system is a relation extractor to populate a target field of the KB with facts extracted from natural language. Relation extraction (RE) is the task of assigning a semantic relationship between (pairs of) entities in text. 
\par There are two categories of RE systems:
\begin{enumerate*}[label=(\roman*)]
\item \emph{closed}-schema IE systems extract relations from a fixed schema or for a closed set of relations, while
\item \emph{open} domain IE systems extract relations defined by arbitrary phrases between arguments.
\end{enumerate*}
We focus on the completion of KBs with a fixed schema, \ie closed IE systems.

\par Effective approaches for closed schema RE apply some form of supervised or semi-supervised learning~\cite{miller2000novel,kambhatla2004combining,boschee2005automatic,jiang2007systematic,sun2011semi,bunescu2005shortest} and generally follow three steps: 
\inlinelist{(\roman*)}{
\item sentences expressing relations are transformed to a data representation, \eg vectors are constructed to be used in feature-based methods, 
\item a binary or multi-class classifier is trained from positive and negative instances, and \item the model is then applied to new or unseen instances.}  \revision{To review the evolution of these and other Natural Language Processing (NLP) techniques, readers can refer to the article by Cambria and White~\cite{cambria2014jumping}.}

Supervised systems are limited by the availability of expensive training data. To counter this problem, the technique of iterative bootstrapping has been proposed~\cite{agichtein2000snowball,gupta2014spied}, in which an initial seed set of known facts is used to learn patterns, which in turn are used to learn new facts and incrementally extend the training set. These bootstrapping approaches suffer from semantic drift and are highly dependent on the initial seed set.

%#\begin{figure}[t!,width=\textwidth]
\begin{figure}[t!]
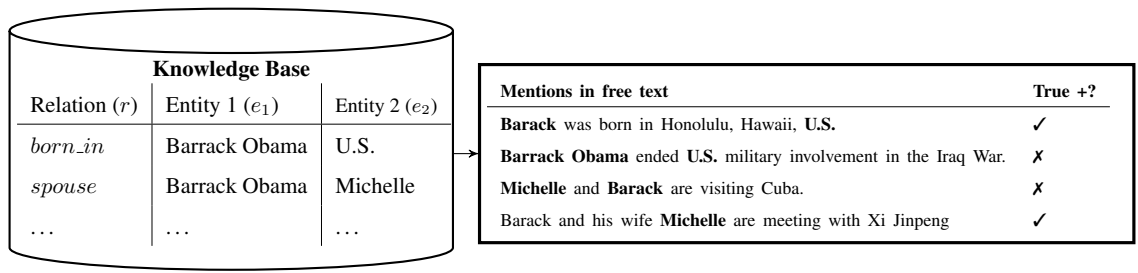

\centering
		\include{figs/ds_example}
	\caption{Illustration of the distant supervision paradigm and errors}
	\label{fig:ds_example}
\end{figure}

When an existing KB is available, a much larger set of known facts can be used to bootstrap training data, a procedure known as Distant Supervision (DS). DS automatically labels its own training data by heuristically aligning facts from a KB with an unlabeled corpus. The KB, written as $D$, can be seen as a collection of relational tables $r(e_1, e_2)$, in which $r\in R$ ($R$ is the set of relation labels), and $<e_1,e_2 >$ is a pair of entities that are known to have relation $r$. The corpus is written as $C$.

The intuition underlying DS is that any sentence in $C$ which mentions the same pair of entities ($e_1$ and $e_2$), expresses a particular relationship $\hat{r}$ between them, which most likely corresponds to the known fact from the KB, $\hat{r}(e_1,e_2)=r(e_1,e_2)$, and thus forms a positive training example for an extractor of relation~$r$. 
DS has been successfully applied in many relation extraction tasks~\cite{surdeanu2014overview,shin2015incremental} as it allows for the creation of large training sets with little or no human effort. 

\par Equally apparent from the above intuition, however, is the danger of finding incorrect examples for the intended relation. The heuristic of accepting each co-occurrence of the entity pair $<e_1,e_2 >$ as a positive training item because of the KB entry $r(e_1, e_2)$, is known to generate noisy training data or false positives~\cite{riedel2010modeling}, \ie two entities co-occurring in text are not guaranteed to express the same relation as the field in the KB they were generated from. The same goes for the generation of negative examples: training data consisting of facts missing from the KB are not guaranteed to be false since a KB in practice is highly incomplete. An illustration of DS generating noisy training data is shown in Figure~\ref{fig:ds_example}.

\par Several strategies have been proposed to reduce this noise. The most prominent is that of latent variable models of the distantly supervised data that make the assumption that a known fact is expressed at least once in the corpus~\cite{riedel2010modeling,hoffmann2011knowledge,surdeanu2012multi}. These methods are cumbersome to train and are sensitive to initialization parameters of the model.

\par An active research direction is the combination of DS with partial supervision, as was proposed in several recent works which differ in the way this supervision is chosen and included. Some focus on active learning, selecting training instances to be labeled according to an uncertainty criterion~\cite{angeli2014combining,surdeanu2014overview}, while others focus on annotations of surface patterns and define rules or guidelines in a semi-supervised learning setting~\cite{pershina2014infusion}. Existing methods for fusion of distant and partial supervision require thousands of annotations and hours of manual labor for minor improvements \revision{(4\% in \Fone for 23,425 annotations~\cite{angeli2014combining} or 2,500 labeled sentences indicating true positives for a 3.9\% gain in \Fone~\cite{pershina2014infusion})}. In this work we start from a distantly supervised training set and show that, using minimal supervision, we can reduce noise in the training data and boost extraction performance. We will demonstrate how only a couple of minutes of annotation time per relation suffices to strongly reduce noise, and obtain significant improvements in precision and recall of the extracted relations.

\par We define the following research questions:
\begin{resq} How can we add supervision most effectively to reduce noise and optimize relation extractors?\end{resq}
\begin{resq} Can we combine semi-supervised learning and dimension reduction techniques to further enhance the quality of the training data and obtain state-of-the-art results using minimal manual supervision?\end{resq}

With the following contributions, we provide answers to these research questions:
\begin{enumerate}
	\item In answer to RQ 1, we demonstrate the effectiveness and efficiency of filtering training data based on high-precision trigger patterns. These are obtained by training initial weak classifiers and manually labeling a small amount of features chosen according to an active learning criterion.
	\item We tackle RQ 2 by proposing a semi-supervised learning technique that allows extending an initial set of high-quality training instances with weakly supervised candidate training items by measuring their similarity in a low-dimensional semantic vector space. This technique is called Semantic Label Propagation.
\item We evaluate our methodology on test data from the English Slot Filling (ESF) task of the Knowledge Base Population at the 2014 Text Analysis Conference (TAC). We compare different methods by using them in an existing KBP system. Our relation extractors attain state-of-the-art effectiveness (a micro averaged \Fone value of 36\%) while relying on a very low manual annotation cost (\ie 5 minutes per relation).
\end{enumerate}

\par In Section~\ref{sec:RW} we give an overview of existing supervised and semi-supervised RE methods and highlight their remaining shortcomings. Section~\ref{sec:system} describes our proposed methodology, with some details on the DS starting point (Section~\ref{sec:ds}), the manual feature annotation approach  (Section~\ref{sec:hcoc}), and the introduction of the semantic label propagation method (Section~\ref{sec:semantic}).
The experimental results are given in Section~\ref{sec:results}, followed by our conclusions in Section~\ref{sec:conclusions}.

\section{Related Work}
\label{sec:RW}

The key idea of our proposed approach is to combine distant supervision with a minimal amount of supervision, 
\ie requiring as few (feature) annotations as possible.
%using minimal amount of (feature) annotations. 
Thus, our work is to be framed in the context of supervised and semi-supervised relation extraction (RE), and related to approaches designed to minimize the annotation cost, \eg active learning. Furthermore, we use compact vector representations carrying semantics, \ie so-called word embeddings. Below, we therefore briefly summarize related work in the areas of
\begin{enumerate*}[label=(\roman*)]
\item supervised RE,
\item semi-supervised RE,
\item active learning, and
\item word embeddings.
\end{enumerate*}

%Our work is related to distant supervision, semi-supervised learning, active learning and distributional semantics.
%\chris{Maybe indicate where (related work in) these areas are discussed in the next subsections? If not, this sole sentence looks a bit forlorn.}

%\chris{Re terminology: maybe revise, and stick to a minimal, consistent set of terms (``instances'', ``facts'', ``mentions'' \ldots)}

\subsection{Supervised Relation Extraction}
Supervised RE methods rely on training data in the form of 
sentences tagged with a label indicating the presence or absence of the considered relation. 
There are three broad classes of supervised RE:
\inlinelist{(\roman*)}{
	\item methods based on manual feature engineering,
	\item kernel based methods, and
	\item convolutional neural nets.
}

\emph{Methods based on feature-engineering}~\cite{jiang2007systematic,mintz2009distant} extract a rich list of manually designed structural, lexical, syntactic and semantic features to represent the given relation mentions. These features are cues for the decision whether the relation is present or not. 
Afterwards a classifier is trained on positive and negative examples. 
In contrast, \emph{kernel based methods}~\cite{Zelenko2002,culotta2004dependency,bunescu2005shortest} represent each relation mention as an object such as an augmented token sequence or a parse tree, and use a carefully designed kernel function, e.g., subsequence kernel or a convolution tree kernel, to calculate their similarity with test patterns. These objects are usually augmented with extra features such as semantic information. 
With the recent success of deep neural networks in NLP, \emph{Convolutional Neural Networks (CNNs)} have emerged as effective relation extractors~\cite{zeng2014relation,xu2015semantic}. CNNs avoid the need for preprocessing and feature design by transforming tokens into dense vectors using embeddings of words. Lexical and sentence-level features are extracted using deep neural nets. Finally, the features are fed into a soft-max classifier to predict the relationship between two marked nouns.
%
% Weggelaten, wordt in introductie al verteld
% \par Focus in our work is on generation of training data, which we will apply in feature based methods. Given a sentence $s$ and two entity mentions $e1$ and $e2$ contained in $s$, a candidate relation mention $r$ from a fixed set of possible relations $R$ is defined as $r = (s, e1, e2)$. The goal is to determine whether $r$ expresses the type defined, and if so, classify it into one of the relation types. RE is then posed as a classification problem and solves it with supervised machine learning classifiers such as Logistic Regression~\cite{jiang2007systematic,mintz2009distant} or Support Vector Machine~\cite{roth2013effective}. Zhou\etal \cite{guodong2005exploring} and Jiang\etal \cite{jiang2007systematic} present a systematic exploration of feature space for RE.

\par Supervised approaches all share the need for training data, which is expensive to obtain. Two common methods have emerged for the generation of large quantities of training data, and both require an initial set of known instances. When this number is initially small, the technique of \emph{bootstrapping} is used. When a very large number of instances is available from an existing knowledge base, \emph{distant supervision} is the preferred technique. Both are briefly discussed below.

\subsubsection{Bootstrapping models for Relation Extraction}
\label{sec:bootstrapping}
When a limited set of labeled instances is available, bootstrapping methods have proven to be effective methods to generate high-precision relation patterns~\cite{agichtein2000snowball,gupta2014spied,ilprints421,Zhang2015128}. The objective of bootstrapping is to expand an initial `seed' set of instances with new relationship instances. Documents are scanned for entities from the seed instances and linguistic patterns connecting them are extracted. Patterns are then ranked according to coverage (recall) and low error rate (precision). Using the top scoring patterns, new seed instances are extracted and the cycle is repeated. 

\par An important step in bootstrapping methods is the calculation of similarity between new patterns and the ones in the seed set. This measure decides whether a new pattern is relation oriented or not, based on the existing set. Systems use measures based on exact matches~\cite{ilprints421}, cosine-similarity~\cite{agichtein2000snowball} or kernels~\cite{Zhang2015128}. A fundamental problem of these methods is semantic drift~\cite{Komachi2008,Curran2007}: bootstrapping, after several iterations, deviates from the semantics of the seed relationship and extracts unrelated instances which in turn generate faulty patterns. This phenomenon gets worse with the number of iterations of the bootstrapping process.

\par Recently, Batista \etal \cite{batista2015} proposed the use of word embeddings for capturing semantic similarity between patterns. Contexts are modeled using linear combinations of the word embeddings and similarity is measured in the resulting space. This approach has shown to reduce semantic drift compared to previous similarity measures.

\subsubsection{Distant Supervision}
Distant Supervision (DS) was first proposed in~\cite{craven1999constructing}, where labeled data was generated by aligning instances from the Yeast Protein Database into research articles to train an extractor. This approach was later applied for training of relation extractors between entities in~\cite{min2013distant}.
%introduce all sentence assumption (without explicitly naming it)
\par 
Automatically gathering training data with DS is governed by the assumption that \textit{all sentences} containing both entities engaged in a reference instance of a particular relation, represent that relation.
Many methods have been proposed to reduce the noise in training sets from DS. In a series of works the labels of DS data are seen as latent variables. 
Riedel \etal \cite{riedel2010modeling} relaxed the strong \textit{all sentences}-assumption and relaxed it to an \textit{at-least-one-sentence}-assumption, creating a Multi-Instance learner. Hoffman \etal\cite{Hoffmann2009} modified this model by allowing entity pairs to express multiple relations, resulting in a Multi-Instance Multi-Label setting (MIML-RE). Surdeanu \etal \cite{surdeanu2012multi} further extended this approach and included a secondary classifier, which jointly modeled all the sentences in texts and all labels in knowledge bases for a given entity pair.
\par Other methods apply heuristics~\cite{intxaurrondo2013removing}, model the training data as a generative process~\cite{alfonseca2012pattern,takamatsu2012reducing} or use a low-rank representation of the feature-label matrix to exploit the underlying semantic correlated information.

\subsection{Semi-supervised Relation Extraction}
Semi-supervised Learning is situated between supervised and unsupervised learning. In addition to unlabeled data, algorithms are provided with some supervised information. The training data comprises labeled instances $X_l=(x_1 \ldots x_l)$ for which labels $Y_l=(y_1 \ldots y_l)$ are provided, and typically a large set of unlabeled ones $X_u=(x_1 \ldots x_u)$.

\par Semi-supervised techniques have been applied to RE on several occasions. Chen \etal \cite{chen2006relation} apply Label Propagation by representing labeled and unlabeled examples as nodes and their similarities as the weights of edges in a graph. In the classification process, the labels of unlabeled examples are then propagated from the labeled to unlabeled instances according to similarity.
Experimental results demonstrate that this graph-based algorithm can outperform SVM in terms of \Fone when very few labeled examples are available. Sun \etal \cite{sun2011semi} show that several different word cluster-based features trained on large corpora can compensate for the sparsity of lexical features and thus improve the RE effectiveness.
%
% An active research direction is the combination of semi-supervision with Distant Supervision (DS).
\par Zhang \etal \cite{zhang2012big} compare DS and complete supervision as training resources but do not attempt to fuse them. They observe that distant supervision systems are often recall gated: to improve distant supervision quality, large input collections are needed. They also report modest improvements by adding crowd-sourced yes/no votes to the training instances. Training instances were selected at random as labeling using active learning criteria did not affect performance significantly.
\par Angeli \etal \cite{angeli2014combining} show that providing a relatively small number of mention-level annotations can improve the accuracy of MIML-RE. They introduce an active learning criterion  for the selection of instances incorporating both the uncertainty and the representativeness, and show that the choice of criterion is important. The MIML-RE model of Surdeanu \etal \cite{surdeanu2012multi} marginally outperforms the Mintz++ baseline using solely distant supervision\revision{:} initialization of the latent variables using labeled data is needed for larger improvements. For this, a total of $10,000$ instances were labeled\revision{, resulting in} a 3\% increase on the micro-{\Fone}. 

Guided Distant Supervision, proposed by Pershina \etal \cite{pershina2014infusion}, incorporates labeled patterns and trigger words to guide MIML-RE during training. They make use of a labeled dataset from TAC KBP to extract training guidelines, which are intended to generalize across many examples.

\subsection{TAC KBP English Slot Filling} 
\revision{
The Knowledge Base Population (KBP) shared task is part of the NIST Text Analysis Conference and aims to evaluate different approaches for discovering facts about entities and expansion of Knowledge Bases. A selection of entities is distributed among participants for which missing facts need to be extracted from a given large collection of news articles and internet fora. Important components of these systems are query expansion, entity linking and relation extractors. Over the years Distant Supervision has become a regular feature of effective systems~\cite{surdeanu2014overview,ji2011knowledge}. Other approaches use hand-coded rules or are based on Question Answering systems~\cite{ji2011knowledge}.
The top performing 2014 KBP ESF system~\cite{angeli2014stanford} uses DS, the manual labeling of $100,000$ features, and is built on DeepDive, a database system allowing users to rapidly construct sophisticated end-to-end knowledge base population techniques~\cite{zhang2015deepdive}. After initial DS, features are manually labeled and only pairs associated with labeled features are used as positive examples. This approach has proven to be very effective but further investigation is needed to reduce the amount of feature labeling. Here, we show how we can strongly reduce this effort while maintaining high precision.%
}

\subsection{Active Learning and Feature Labeling}
Active learning is used to reduce the amount of supervision required for effective learning. The most popular form of active learning is based on iteratively requiring manual labels for the most informative instances, an approach called uncertainty sampling.
In relation extraction, typical approaches include query-by-committee~\cite{angeli2014combining,seung1992query} and cluster-based sampling~\cite{sterckx2014using}.
While the focus in RE has been on labeling relation instances, alternative methods have been proposed in other tasks in which features (\eg patterns, or the occurrence of terms) are labeled as opposed to instances~\cite{druck2009active,Attenberg10aunified}, resulting in a higher performance for less supervision.

Getting positive examples for certain relations can be hard, especially when training data is weakly supervised. Standard uncertainty sampling is ineffective in this case: it is likely that a feature or instance has a low certainty score because it does not carry much discriminative information about the classes. Assigning labels to the most certain features has much greater impact on the classifier and can remove the principle sources of noise. This approach has been coined as Feature Certainty~\cite{Attenberg10aunified}, and we show that this approach is especially effective in DS for features that generalize across many training instances.

\subsection{Distributional Semantics}
The Distributional Hypothesis~\cite{harris54} states that words that tend to occur in similar contexts are likely to have similar meanings. 
Representations of words as dense vectors (as opposed to the standard one-hot vectors), called word embeddings, exploit this hypothesis and are trained from large amounts of unlabeled data on predicting their context.
Representations for words will be similar to those of related words, allowing the model to generalize better to unseen events.  
% The similarity between representations of terms not explicitly present in the labeled training data, with those of terms related in meaning that do appear in it, allow for a better generalization to unseen events.
The resulting vector space is also called a \textit{vector model of meaning}~\cite{martin2015speech}.
Common methods for generating very dense, short vectors use dimensionality reduction techniques (\eg singular value decomposition) or neural nets to create so-called word embeddings. Word embeddings have proven to be beneficial for many Natural Language Processing tasks including POS-tagging, machine translation and semantic role labeling.
Common unsupervised word embedding algorithms include \textit{Word2Vec}~\cite{mikolov2013} and \textit{GloVe}~\cite{pennington2014}. These models are inspired by neural networks and are trained using stochastic gradient training.

\par While much research has been directed at ways of constructing distributional representations of individual words, for example co-occurrence based representations and word embeddings, there has been far less consensus regarding the representation of larger constructions such as phrases and sentences from these representations. Blacoe \etal ~\cite{blacoe2012} show that, for short phrases, a simple composition like addition or multiplication of the distributional word representations is competitive with more complex supervised models such as recursive neural networks (RNNs).

\section{Labeling Strategy for Noise Reduction}
\label{sec:system}
In this section we introduce our strategy to combine distantly supervised training data with minimal amounts of supervision. Shortly stated, we designed our labeling strategy such as to \emph{minimize the amount of false positive instances or noise while maintaining the diversity of relation expressions generated by DS}.

% \begin{center}
%\textit{our labeling strategy is to minimize the amount of false positive instances or noise while maintaining the diversity of relation expressions generated by DS.}
% \end{center}

\par We perform a highly selective form of noise reduction starting from a fully distantly supervised relation extractor, described in Section~\ref{sec:ds}, and use the feature weights of this initial extractor to guide manual supervision in the feature space.
Various questions arise from this. When do we over-constrain the original training set generated by distant supervision? What is the trade-of between the application of distant supervision with highly diverse labeled instances, and the constraining approach of labeling features, with a highly accurate yet restricted set of training data?
This is discussed in detail in Sections~\ref{sec:hcoc} and~\ref{sec:semantic}.
%\par When labeling features an active learning criterion is applied. 

\par
\begin{figure}%[t,width=\textwidth]
\centering
        \includegraphics[width=\textwidth]{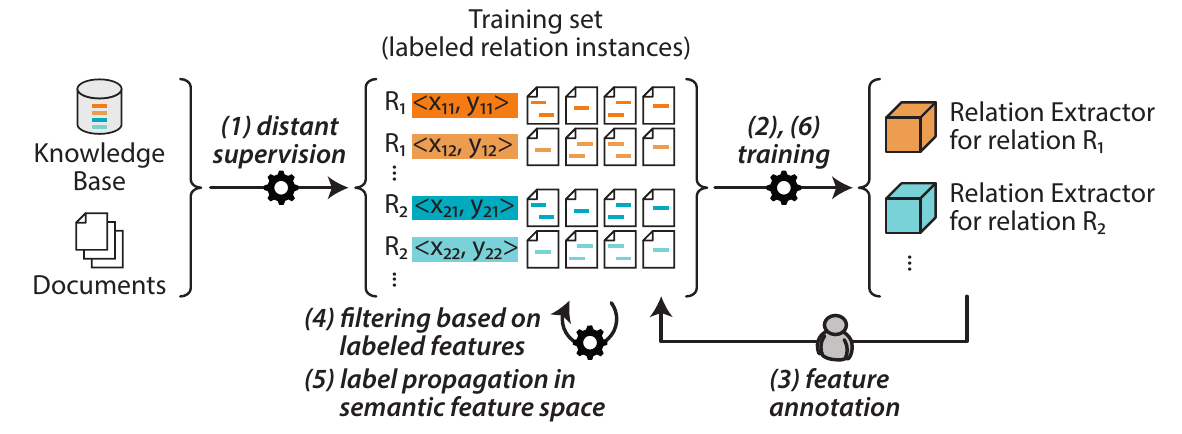}
	\caption{Workflow Overview. \revision{Note that only step (3) involves human annotation effort.}}
	\label{fig:sys}
\end{figure}

 More concretely, our approach is depicted in Figure~\ref{fig:sys}, and comprises the following steps:
%An overview of this workflow is depicted in Figure~\ref{fig:sys}.

\begin{enumerate}[(1)]
\item An existing KB is used to generate distantly supervised training instances by matching its facts with sentences from a large text corpus. We discuss the characteristics of this weakly labeled training set as well as the features extracted from each sentence (see Section~\ref{sec:ds}).
\item An initial relation extractor is trained using the noisy training data generated in step (1).
\item Confident positive features learned by this initial classifier are presented to an annotator with knowledge of the semantics of the relation and labeled as true positive or false positive. Feature confidence is quantified with an active learning criterion.
\item The collection of training instances is filtered according to the labeled features and a second classifier is trained. This framework, in which we combine supervision and DS, is explained in Section~\ref{sec:hcoc}.
\item In a \textit{semi-supervised} step, the filtered distantly supervised training data is added to training data by propagating labels from labeled features to distantly supervised instances based on similarity in a semantic vector space of reduced dimension. The technique is presented in~\ref{sec:semantic} as Semantic Label Propagation. 
\item A final relation extractor is trained from the augmented training set. We evaluate and discuss results of the proposed techniques in Section~\ref{sec:results}. 
\end{enumerate}

% \tikzsetnextfilename{system}

\subsection{Distantly Supervised Training Data}
\label{sec:ds}
The English Gigaword corpus~\cite{graff2003english} is used as unlabeled text collection to generate relation mentions. The corpus consists of 1.8 million news articles published between January 1987 and June 2007. Articles are first preprocessed using different components of the Stanford CoreNLP toolkit~\cite{manning2014stanford}, including sentence segmentation, tokenizing, POS-tagging, Named Entity Recognition, and clustering noun phrases which refer to the same entity. 
\par As KB we use a snapshot of Freebase (now Wikidata) from May 2013. The relation schema of Freebase is mapped to that used for evaluation, the NIST TAC KBP ESF Task, which defines 41 relations, including 25 relations with a person as subject entity and 16 with organizations as subject. 26 relations require objects or fillers that are themselves named entities (\eg Scranton as place of birth of Joe Biden), whereas others require string-values (\eg profession (Senator, Teacher, \ldots), cause of death (Cancer, Car Accident,\ldots)).

\par We perform weak Entity Linking between Freebase Entities and textual mentions using simple string matching.
\revision{ We reduce the effect of faulty entity links by thresholding the amount of training data per subject entity~\cite{UGent_TAC_2014}. Most frequently occurring entities from the training data (\eg John Smith, Robert Johnson, \ldots) are often most ambiguous, hard to link to a KB and thus result in noisy training data. Thresholding the amount of training data per entity also prevents the classifier from overfitting on several, popular entities. The reason for that is our observation that training data was initially skewed towards several entities frequently occurring in news articles, like Barack Obama or the United Nations, resulting in over-classifying professions of persons as president or seeing countries as members of the organization.
}
% \todo[inline]{@Lucas: 2nd sentence of revision above is malformed... no main verb? Missing part of sentence?}

\par For each generated pair of mentions, we compute various lexical, syntactic and semantic features. Table~\ref{tab:features} shows an overview of all the features applied for the relation classification. We use these features in a binary Logistic Regression classifier. Features are illustrated for an example relation-instance $<$Ray Young, General Motors$>$ and the sentence \textit{``Ray Young, the chief financial officer of General Motors, said GM could not bail out Delphi''}.

\begin{table}[htbp]
\footnotesize
\caption{Overview of different features used for classification for the sentence ``Ray Young, the chief financial officer of General Motors, said GM could not bail out Delphi''.}
	\centering
  \label{tab:features}
\begin{tabular}{ccc}
\toprule
\multicolumn{1}{c}{\textbf{Feature}} & \textbf{Description} & \textbf{Example Feature Value} \\ 
\midrule
\multicolumn{1}{c}{\multirow{6}{*}{\makecell{Dependency\\tree}}} 	 & \makecell{Shortest path connecting the two names in\\the dependency parsing tree coupled with\\entity types of the two names} & \makecell{PERSON$\leftarrow$appos$\leftarrow$officer \\$\rightarrow$ prep\_of$\rightarrow$ ORGANIZATION } \\
\cmidrule{2-3}
\multicolumn{1}{c}{}  & The head word for name one & said \\ \cmidrule{2-3}
\multicolumn{1}{c}{}  & The head word for name two & officer \\ \cmidrule{2-3}
\multicolumn{1}{c}{}  & Whether \textit{1dh} is the same as \textit{e2dh} & false \\ \cmidrule{2-3}
\multicolumn{1}{c}{}  & The dependent word for name one & officer \\ \cmidrule{2-3}
\multicolumn{1}{c}{}  & The dependent word for name two & nil \\
\midrule
\multicolumn{1}{c}{\multirow{9}{*}{\makecell{Token\\sequence\\features}}}  & The middle token sequence pattern & , the chief financial officer of \\ \cmidrule{2-3}
\multicolumn{1}{c}{}  & Number of tokens between the two names & 6 \\ \cmidrule{2-3}
\multicolumn{1}{c}{}  & First token in between &, \\ \cmidrule{2-3}
\multicolumn{1}{c}{}  & Last token in between & of \\ \cmidrule{2-3}
\multicolumn{1}{c}{}  & Other tokens in between & \{the, chief, financial, officer\}
 \\ \cline{2-3}
\multicolumn{1}{c}{}  & First token before the first name & nil \\ \cmidrule{2-3}
\multicolumn{1}{c}{}  & Second token before the first name & nil \\ \cmidrule{2-3}
\multicolumn{1}{c}{}  & First token after the second name & , \\ \cmidrule{2-3}
\multicolumn{1}{c}{}  & Second token after the second name & said \\
\midrule 
\multicolumn{1}{c}{\multirow{6}{*}{\makecell{Entity\\features}}}  & String of name one & Ray\_Young \\ \cmidrule{2-3}
\multicolumn{1}{c}{}  & String of name two & General\_Motors \\ \cline{2-3}
\multicolumn{1}{c}{}  & Conjunction of \textit{e1} and \textit{e2} & Ray\_Young--General\_Motors \\ \cmidrule{2-3}
\multicolumn{1}{c}{}  & Entity type of name one & PERSON \\ \cmidrule{2-3}
\multicolumn{1}{c}{}  & Entity type of name two & ORGANIZATION \\ \cmidrule{2-3}
\multicolumn{1}{c}{}  & Conjunction of \textit{et1} and \textit{et2} & PERSON--ORGANIZATION \\
\midrule
\multicolumn{1}{c}{\makecell{Semantic\\feature}}  & Title in between & True \\
\midrule
\multicolumn{1}{c}{\makecell{Order\\feature}} & \makecell{1 if name one comes before name two;\\2 otherwise.} & 1\\ 
\midrule
\multicolumn{1}{c}{Parse Tree}  & \makecell{POS-tags on the path connecting\\the two names} & \makecell{NNP$\rightarrow$DT$\rightarrow$JJ$\rightarrow$JJ\\$\rightarrow$NN$\rightarrow$IN$\rightarrow$NNP }\\ 
\bottomrule
\end{tabular} 
\label{tab:features}
\end{table}

\par For each relation $R_i$, we generate a set of (noisy) positive examples denoted as $R_i^+$ and defined as $$R^+_i =
\{\ (m_1, m_2) \;|\; R_i(e_1, e_2) ∧ EL(e_1, m_1) ∧ EL(e_2, m_2)\ \}$$
with $e_1$ and $e_2$ being subject and object entities from the KB and $EL(e_1, m_1)$ being the entity $e_1$ linked to mention $m_1$ in the text. 
As in previous work~\cite{Hoffmann2009,Mintz2009}, we impose the constraint that both entity mentions $(m_1, m_2) \in R^+_i$ are contained in the same sentence. To generate negative examples for each relation, we sample instances from co-occurring entities for which the relation is not present in the KB.

\par \revision{We measured the amount of noise, \ie false positives, in the training set of positive DS instances, for a selection of 15 relations: we manually verified 2,000 randomly chosen instances (that DS found as supposedly positive examples) for each of these relations. Table~\ref{tab:ratio} shows the percentage of true positives among these 200 instances for each of these relations, which strongly varies among relations, ranging from 10\% to 90\%.}

% \todo[inline]{@Lucas: Maybe briefly indicate in Table~\ref{tab:ratio}'s caption how exactly the fraction of true positives is ``estimated''?}

\begin{table}[htbp]
\small
\caption{Training Data. Fractions of true positives are estimated \revision{from the training data by manually labeling} a sample of 2,000 instances per relation \revision{that DS indicated as positive examples}.}
	\centering
  \label{tab:ratio}
  \begin{tabular}{lccccc}
    \toprule
%\textbf{Relation} & \pbox{3cm}{ \revision{\textbf{Estimated Fraction}}\\ \revision{\textbf{of True Positives}}} & \pbox{3cm}{\revision{\textbf{Positively}} \\ \revision{\textbf{Labeled SDPs}}} & \pbox{3cm}{\textbf{Remaining Training}\\ \textbf{data after Filtering}} & \pbox{3cm}{\revision{\textbf{Initial Number of}} \\ \revision{\textbf{True Positives}}}  \\ 
\multirow{2}{*}{\textbf{Relation}} & \revision{\textbf{Estimated Fraction}} & \revision{\textbf{Positively}} & \textbf{Remaining Training} & \revision{\textbf{Initial Number of}} \\
 & \revision{\textbf{of True Positives}} & \revision{\textbf{Labeled SDPs}} & \textbf{Data after Filtering} & \revision{\textbf{True Positives}} \\
\midrule
per:title &  85.1\% & 157 & 26.2\% & 369,079\\
org:top\_members\_employees &  71.7\%& 236 & 16.7\% & 93,900 \\
per:employee\_or\_member\_of &  87.8\% & 256 & 16.5\%& 260,785 \\
per:age &  62.4\% & 79 & 52.2\%  &  58,980  \\
per:origin &  85.2\%& 116 & 11.9\% & 1,555,478\\
per:countries\_of\_residence & 55.6\%& \revision{65} & 8.4\%&493,064 \\
per:charges &  59.4\%& 122 & 21.5\% & 17,639\\
per:cities\_of\_residence &  11.7\%& 96 & 7.4\%& 370,153 \\
per:cause\_of\_death&  51.9\%& 97 & 29.4\% & 31,386 \\
per:spouse &  63.2\%& 124 & 12.1\% & 172,874 \\
per:city\_of\_death &  19.9\% & 92 & 5.6\% & 125,333\\
org:country\_of\_headquarters &  10.8\%& 92 & 13.4\% &13,435 \\
per:country\_of\_death &  77.6\% & 70 & 16.5\% & 128,773 \\
org:city\_of\_headquarters &  56.5\%& 67 & 42.7\% &36,238 \\
% org:alternate\_names &  49.6\%& 20 & 43.4\% &109,582 \\
org:founded\_by &  13.3\% & 85 & 22.7\% &318,991 \\
% per:country\_of\_birth & 99.5\%& 100 & 1.9\% &1,491,144 \\
% per:schools\_attended &  62.6\% & 58 & 7.4\% & 31,852\\
\bottomrule
\end{tabular}
\end{table}
\normalsize

\subsection{Labeling High Confidence Shortest Dependency Paths}
\label{sec:hcoc}
\begin{figure}
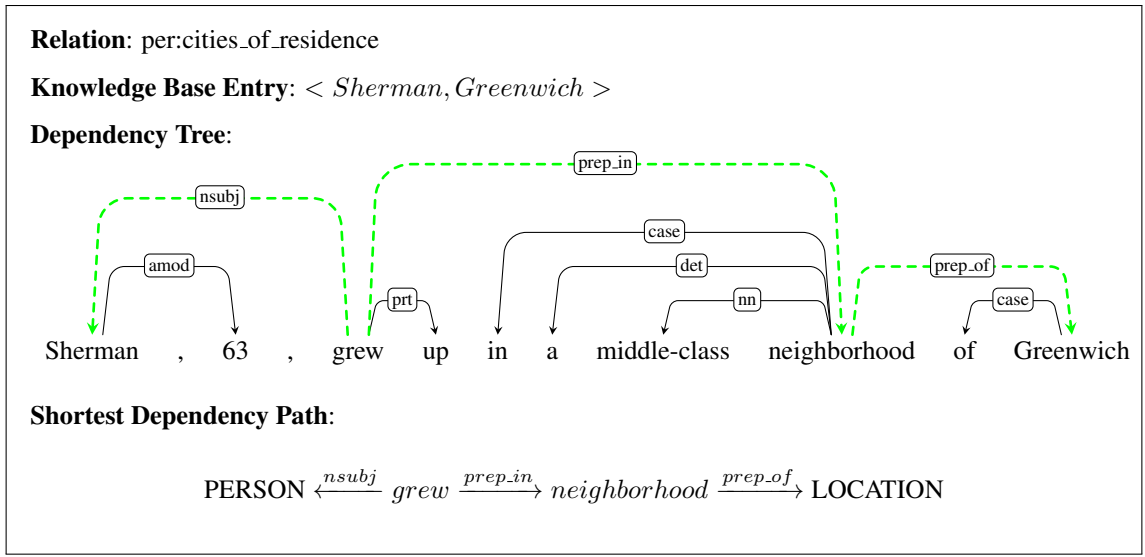

\begin{framed}
\textbf{Relation}: per:cities\_of\_residence \newline
\textbf{Knowledge Base Entry}:  $<Sherman, Greenwich>$ \newline
\textbf{Dependency Tree}:\newline
% In the document:
\begin{dependency}
   \begin{deptext}[column sep=1em]
   Sherman \&, \& 63\& ,\& grew\& up \& in \& a\& middle-class\& neighborhood \& of\& Greenwich \&. \& \\
   \end{deptext}
  % \deproot{5}{ROOT}
   \depedge[edge style={green,dashed,line width=1.0pt}]{5}{1}{nsubj}
   \depedge{5}{6}{prt}
   \depedge[edge style={green,dashed,line width=1.0pt}]{5}{10}{prep\_in}
   
   \depedge{1}{3}{amod}
   
   \depedge[edge style={green,dashed,line width=1.0pt}]{10}{12}{prep\_of}
   \depedge{10}{7}{case}
   \depedge{10}{8}{det}
   \depedge{10}{9}{nn}
   
   \depedge{12}{11}{case}
\end{dependency}\newline
\textbf{Shortest Dependency Path}:
$$\text{PERSON}\xleftarrow{nsubj} grew \xrightarrow{prep\_in}neighborhood \xrightarrow{prep\_of}\text{LOCATION}  $$
\end{framed}

	\caption{Dependency tree feature}
	\label{fig:dtree_example}
\end{figure}

This section describes the manual feature labeling step that allows transforming a full DS training set into a strongly reduced yet highly accurate training set, based on manual feature labeling.
We focus on a particular kind of feature, \ie a relation's Shortests Dependency Path (SDP).
Dependency paths have empirically been proven to be very informative for relation extraction\delete{,}: their capability in capturing a lot of information is evidenced by a systematic comparison in effectiveness of different kernel methods~\cite{stevenson2006comparing} or as features in feature-based systems~\cite{jiang2007systematic}. 
This was originally proposed by Bunescu \etal~\cite{bunescu2005shortest}, who claimed that the relation expressed by a sentence is often captured in the \textit{shortest} path connecting the entities in the dependency graph. 
Figure~\ref{fig:dtree_example} shows an example of an SDP for a sentence expressing a relation between a person and a city of residence. 

% % \tikzsetnextfilename{conf_sparse}
% \begin{figure}[!tbp]
%   \centering
%   \begin{minipage}[b]{0.45\textwidth}
% 		\include{figs/cumprob}
% 	\caption{Occurrence frequency of dependency paths, ranked from most to least frequent, for example relations.}
% 	\label{fig:sparse}
%   \end{minipage}
%   \hfill
%   \begin{minipage}[b]{0.45\textwidth}
%     		\include{figs/conf}
% %  		\includegraphics[width=10cm]{figs/per_3Acity_of_death.png}
% 	\caption{Confidence of dependency paths, ranked from most to least confident, for example relations, with indication of true positives.}
% 	\label{fig:conf}
%   \end{minipage}
% \end{figure}

\begin{figure}[!tbp]
  \centering
  \subfloat[]{\label{fig:sparse}
	  \input{figs/cumprob}
  }
  \quad
  \subfloat[]{\label{fig:conf}
	  \input{figs/conf}
  }
  \caption{Illustration of frequency and confidence of dependency paths for example relations. 
  (a) Occurrence frequency, ranked from highest to lowest, and (b) confidence $C$ of dependency paths (eq.~\ref{eq:confidence}), ranked from highest to lowest, with indication of true positives.}
\end{figure}
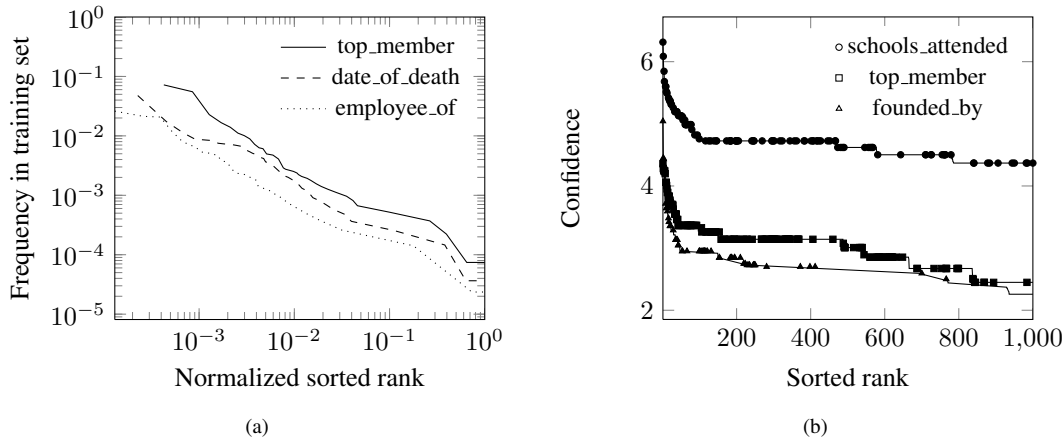

% \begin{figure}
% \centering
% 		\include{figs/cumprob}
% 	\caption{Feature Sparsity}
% 	\label{fig:sparse}
% \end{figure}

% \begin{figure}
% \centering
% 		\include{figs/pathlengths}
% 	\caption{Average Length}
% 	\label{fig:lens}
% \end{figure}

\par As shown in Table~\ref{tab:ratio}, the fraction of false positive items among all weakly supervised instances can be very large. Labeling features based on the standard active learning approach of uncertainty sampling is ineffective in our case since it is likely that a feature or instance has a low certainty score simply because not much discriminative information about the classes is carried. Annotating many such instances would be a waste of effort. Assigning labels to the most certain features has much greater impact on the classifier and can remove the principal sources of noise. This approach is called \textit{Feature Certainty Sampling}~\cite{Attenberg10aunified}.
It is intuitively an attractive method, as the goal is to reduce the most influential sources of noise as fast as possible. For example for the relation \textit{founded\_by} there are many persons that founded the company who are also \textit{top\_members}, leading to instances that we wish to remove when cleaning up the training data for the relation \textit{founded\_by}. %After the feature-labeling stage we further increase the quality of our training set by propagating labels from labeled features to remaining distantly supervised features. 

SDPs offer all the information needed to assess the relationship validity of the training instances, are easily labeled, and generalize over a considerable fraction of the training set as opposed to many of the feature-unigrams which remain ambiguous in many cases. We implement the feature certainty idea by ranking SDP features according to the odds that when a particular SDP occurs, it corresponds to a valid relation instance. This corresponds to ranking by the following quantity, which we call the considered SDP's confidence
\begin{equation} 
\text{Confidence}(\textit{SDP}) = \frac{P(+\vert\textit{SDP})}{P(-\vert\textit{SDP})}. \label{eq:confidence}
\end{equation}
It can be directly estimated from the original distant supervision training set, based on each SDP feature's (smoothed) occurrence frequencies among the positive and negative distantly supervised instances. In particular, $P(+\vert \textit{SDP})$ indicates the SPD's fraction of occurrences among the positive training data and $P(-\vert\textit{SDP})$) among the negative.

%\tocheck{It} can be estimated for each \tocheck{SDP} feature from the original distant supervision training set, based on  \tocheck{its} (smoothed) occurrence frequencies among the positive and negative instances \revision{(where $P(\textit{+\vert SDP})$ indicates the fraction of \tocheck{the SDP in the } SDPs which are labeled as positive trainingdata and similarly for $P(\textit{SDP}|-)$)}, \del{as the parameters of a Naive Bayes classifier}.  \thomasq{dat laatste zou 'k weglaten}

% \todo[inline]{@Lucas: shouldn't you more explicitly define ``$P(\textit{SDP}|+)$'', esp.\ as what the condition ``+'' means exactly? Just a parenthesis could suffice: ``\ldots (where $P(\textit{SDP}|+)$ represents \ldots, and similarly for $P(\textit{SDP}|-)$''.}

\par All dependency paths are ranked from most to least confident and the top-$k$ are assigned to a human annotator to select the true positive SDPs. The annotator is asked to select only the patterns which unambiguously express the relation. That is, a pattern is accepted only if the annotator judges it a sufficient condition for that relation.
The annotator is provided with several complete sentences containing the dependency path to this cause. When the SDP does not include any verbs, \eg when entities are both part of the same Noun Phrase like ``Microsoft CEO Bill Gates", all words between the subject and object are included and the complete path is added to the filter set. In our experiments, we restrict the time of SDP annotations to a limited effort of 5 minutes for each relation. On average our expert annotator was able to label around 250 SDPs per relation this way. The ease of annotating SDPs becomes apparent when comparing with annotating random relation instances, which they managed to do at a rate of only 100 in the same period of time. 
\revision{Section~\ref{sec:annotation_meths} provides further details on the different annotation methodologies for the experiments.}

The motivation behind limiting the annotation time per relation to only a few hundred patterns comes from the following analysis. 
First of all, a small subset of all different patterns are responsible for the majority of relation instances in the DS training set. In fact, the sparsity of Distantly Supervised training data becomes apparent when extracting all SDPs for each fact in the KB in one pass over the corpus. Figure~\ref{fig:sparse} shows the approximately Zipfian distribution of the frequency of the dependency paths generated by DS in the positively labeled training set for several example relations.  
The abscis shows the rank of dependency paths for various relations, sorted from most to least frequent, normalized by the total number of paths for the respective relations (to allow visualization on the same graph).
In line with our goal of getting a highly accurate training set with the largest sources of noise filtered away at a low annotation cost, we need to focus on capturing those top most frequent patterns. 
Secondly, we noticed that beyond the first few hundred most confident SDPs, which take around 5 minutes to annotate, further true positives tend to occur less frequently. Annotating many more SDPs would only marginally increase the diversity in the training set, at a rapidly increasing annotation cost. 
Figure~\ref{fig:conf} illustrates the occurrence of true positive patterns for decreasing confidence scores. For several example relations, the figure shows the true positive patterns as markers on the confidence distribution of the $1,000$ most confident SDPs. 
%The rate of true positive SDPs declines after labeling \red{a few hundred} of the most confident patterns and true positives occur more sparsely.
% \todo[inline]{@Lucas: For those figs.~\ref{fig:sparse} and \ref{fig:conf}: would it be possible to have the legend in the same order (top to bottom) as the corresponding curves? Esp.\ for fig.\ \ref{fig:conf}, the markers are not so easily distinguishable\ldots}

%\par Figure~\ref{fig:lens} shows the correlation between length of the dependency path and corresponding frequency in the training set. This shows that confident patterns are short, include on average only one or two words, and offer a condensed representation of the relation and components that are strong indicators of the relation. Frequently occuring words on these paths have been coined as \textit{triggerwords}~\cite{zeng2014relation,Zhang2015278} in literature.

\par Finally, using the manually selected set of SDPs, the complete training set is filtered by enforcing that one of these SDPs be present in the feature set of the instance. We include all mention pairs associated with that feature as positive examples of the considered relation. The classifier trained on the resulting training set is intuitively of high precision but doesn't generalize well to unseen phrase constructions. Note that the classifier is quite different from a regular pattern based relation extractor. Although all training instances satisfy at least one of the accepted SDPs, the classifier itself is trained on a  set of features including, but not restricted to, these SDPs (see Table~\ref{tab:features}).
%the feature-based approach allows \red{for some generalization}: the use of SDP features in combination with others (see Table~\ref{tab:}) \red{leads directs the extractor in properly weighting} valuable lexical and syntactic features. 
Still, most of the benefits of DS are lost by having the selection of training instances governed by a limited set of patterns.

\begin{table}[t!]
\caption{ Examples of top-ranked patterns }
	\centering
    \small
  \label{tab:Table 1}
  \begin{tabular}{llc}
    \toprule
\textbf{Relation} & \textbf{Top SDP} &  \textbf{Assessment}\\ 
    \midrule
top\_members\_employees & $\text{PER}\xleftarrow{appos} executive \xrightarrow{prep\_of}\text{ORG}$
 &  \ding{51} \\ 
 & $\text{PER}\xleftarrow{appos}chairman\xrightarrow{appos}\text{ORG}$
 &  \ding{51} \\ 
 & $\text{ORG}\xleftarrow{nn}founder\xrightarrow{prep\_of}\text{PER}$
 &  \ding{55} \\ 
     \midrule
children & $\text{PER-2}\xleftarrow{appos} son \xrightarrow{prep\_of}\text{PER-1}$
 &  \ding{51} \\ 
 
 & $\text{PER-1}\xleftarrow{appos}father\xrightarrow{prep\_of}\text{PER-2}$
 &  \ding{51} \\ 
 
 & $\text{PER-2}\xleftarrow{nn}grandson\xrightarrow{prep\_of}\text{PER-1}$
 &  \ding{55} \\ 
      \midrule
city\_of\_birth & $\text{PER}\xleftarrow{rcmod} born \xrightarrow{prep\_in}\text{LOC}$
 &  \ding{51} \\ 
 
 & $\text{PER}\xleftarrow{nsubj}mayor\xrightarrow{prep\_of}\text{LOC}$
 &  \ding{55} \\ 
 
 & $\text{PER}\xleftarrow{appos}historian\xrightarrow{prep\_from}\text{LOC}$
 &  \ding{55} \\ 
 \midrule
schools\_attended & $\text{PER}\xleftarrow{nsubj} graduated \xrightarrow{prep\_from}\text{ORG}$
 &  \ding{51} \\ 
 
 & $\text{PER}\xleftarrow{dep}student\xrightarrow{prep\_at}\text{ORG}$
 &  \ding{51} \\ 
 
 & $\text{PER}\xleftarrow{appos}teacher\xrightarrow{prep\_at}\text{ORG}$
 &  \ding{55} \\ 
  \midrule
(org:)parents & $\text{ORG-2}\xleftarrow{appos} subsidiary \xrightarrow{prep\_of}\text{ORG-1}$
 &  \ding{51} \\ 
 
 & $\text{ORG-1}\xleftarrow{appos}division\xrightarrow{prep\_of}\text{ORG-2}$
 &  \ding{51} \\ 
 
 & $\text{ORG-2}\xleftarrow{prep\_to}shareholder\xrightarrow{dep}\text{ORG-1}$
 &  \ding{55} \\ 
 \bottomrule
\end{tabular}
\label{tab:confidencepatterns}
\end{table}

% After filtering the initial DS training set for previously discussed patterns, a fraction is left on which a classifier of high precision is trained. \thomas{I would leave out the sentence 'After filtering... is trained'; pref. paragraph already explains that.}
The fourth column of Table~\ref{tab:ratio} lists the fraction of training data remaining after filtering out all patterns apart from those classified as indicative of the relation at hand. The amount of training data remaining after this filtering step strongly depends on the specific relation, varying from 5\% to more than half of the original training set. Yet on the whole, the filtering results in a strong reduction of the purely DS-based training data, often removing much more than the actual fraction of noise (column 2). For example, for the relation \textit{per:employee\_or\_member\_of}, \revision{we note only $100\% - 87.8\% = 12.2\%$ false positives}, but the manual filtering leads to discarding 83.5\% of the DS instances.
% \todo[inline]{@Lucas: I don't see a row in Table \ref{tab:ratio} with ``less than 1\%'' training data remaining after filtering?}

\par The strategy described in the previous paragraphs is related to the \textit{guidelines} strategy from Pershina \etal~\cite{pershina2014infusion} (without the MIML model) in labeling features, but it differs in some essential aspects. Instead of needing a fully annotated corpus to do so, we rank and label features entirely based on distant supervision. Labeling features based on a fully labeled set ignores the variety of DS and risks being biased towards the smaller set of labeled instances. Also, no active learning criteria were applied when choosing which features to label, making the process even more efficient.
% \todo[inline]{@Lucas: Reference \cite{pershina2014infusion} has no info as to where this appeared? ACL?}
% \todo[inline]{@Lucas: I also find the ``no selection criteria were applied'' bit a bit vague/unclear: How exactly did \cite{pershina2014infusion}(?) apply further selection criteria, i.e., of what nature where these criteria? It would be nice to at least give a hint. Now it remains mysterious.}

\subsection{Noise Reduction using Semantic Label Propagation}
\label{sec:semantic}
% \begin{figure}
% \centering
% 		\include{figs/semantic}
% 	\caption{Illustration of Semantic Label Propagation in 2 dimensions (on artificial data). (a) Original DS training instances; (b) strongly reduced training set after filtering based on feature annotations; (c) training set after Semantic Label Propagation.}
% 	\label{fig:ssl}
% \end{figure}

% When confining the training set to confident SDP's, the diversity generated by DS is discarded. Many valid training instances are discarded for not matching any of the high precision SDP's. Due to the sparsity of DS data, labeling all patterns or features is not feasible.
% To expand the initial training set, a second stage of relabeling is needed. 
% \par Semi-supervised learning (SSL) is a strategy to expand the labeled data, instead of a set of fully unlabeled trainingdata as in most cases of SSL, our filtered training set is already weakly labeled from DS.  

If we strictly follow the approach proposed in Section~\ref{sec:hcoc} and only retain DS training instances that satisfy an accepted SDP,
%those SDPs judged as indicative of the considered relationship, 
an important advantage of DS is lost, namely its potential of reaching high recall. If we limit the feature annotation effort, we risk losing highly valuable SDPs. To counteract this effect, we introduce a second (re)labeling stage, adopting a semi-supervised learning (SSL) strategy to expand the training set. This is done by again adding some instances from the set of previously discarded DS instances with SDPs not matching any of the manually labeled patterns.
%We start from the weakly labeled DS training set (as opposed to fully unlabeled training data as in most cases of SSL) \ie we look into the set of discarded instances and add a subset of them to the training set with a criterion based on a similarity measure to the initial seed patterns.
%
%\par Patterns such as linked chains have not been used by semi-supervised approaches to pattern learning. 
We rely on the basic SSL approach of self-training by propagating labels from known instances to the nearest neighboring unlabeled instances. This algorithm requires a method of determining the distance between labeled and unlabeled instances.
Dangers of self-training include the failure to expand beyond the initial training data or the introduction of errors into the labeled data. In order to avoid an overly strong focus on the filtered training data, we use low-dimensional vector representations of words, also called word embeddings. 
%apply the dimension reduction technique of word embeddings.
\par Word embeddings allow for a relaxed semantic matching between the labeled seed patterns and the remaining weakly labeled patterns. As shown by Sterckx \etal~\cite{sterckx2014using}, representing small phrases by summing each individual word’s embedding leads to semantic representations of small phrases that are meaningful for the goal of relation extraction. 

\revision{
We represent each relation instance by a single vector by first removing stop-words and averaging the embeddings of the words on the dependency path. For example, consider the sentence:
% $$ \text{\textbf{Vincent Willem van Gogh}\textit{ was born in }\textbf{Groot-Zundert}} $$
% which has the following SDP,
% $$\text{PER}\xleftarrow{appos}born\xrightarrow{prep}in\xrightarrow{pobj}\text{LOC}. $$
%
$$\text{\textbf{Geagea} on Friday for the first time addressed the court judging him for \textbf{murder} charges.}$$
which has the following SDP,
$$\text{PER}\xleftarrow{nsubj}addressed\xrightarrow{dobj}court\xrightarrow{vmod}judging\xrightarrow{prep\_for}charges\xrightarrow{nn}\text{Criminal\_Charge}$$
Its low-dimensional representation $\vec{C}$ is hence generated as
% \begin{equation} \vec{C} = \frac{E(``\text{born}") + E(``\text{in}")}{2}, \label{eq:lowdim}\end{equation}
\begin{equation} \vec{C} = \frac{E(``\text{addressed}") + E(``\text{court}")+ E(``\text{judging}")+ E(``\text{charges}") }{4}, \label{eq:lowdim}\end{equation}
}
with $E(x)$ the word embedding of word $x$. The similarity between a labeled pattern $\vec{C}_t$ and a weakly labeled pattern $\vec{C}_{DS}$ is then measured using cosine similarity between the vector representations.

\begin{equation} Sim(\vec{C}_t, \vec{C}_{DS} ) = \frac{\vec{C}_t . \vec{C}_{DS}}{|\vec{C}_t| . |\vec{C}_{DS}|} \label{eq:lowdim}\end{equation}
In the special case that no verbs occur between two entities, all the words between the two entities are used to build the representations for the context vector.

\par Using these low-dimensional continuous representations of patterns, we can calculate similarities between longer, less frequently occurring patterns in the training data and the patterns from the initial seed set which are the most frequently occurring ones. 
We can now increase recall by adding similar but less frequent patterns.
 
%of frequently occurring and manually labeled patterns, we increase recall by adding patterns with representations having representations near the seed set. 
More specifically, we calculate the similarity of the average vector of the labeled patterns (as in the Rocchio classifier type of self-training) with each of the remaining patterns in the DS set and \revision{extend the training data with the patterns that have a sufficiently high similarity with the labeled ones}. We call this technique \textit{Semantic Label Propagation}.
% \revision{The process of Semantic Label Propagation is visualized using artificial data in Figure~\ref{fig:ssl}. A first stage depicts all SDPs used for the initially distantly supervised training data. In a second stage several high precision SDPs are labeled by an annotator and false positives are filtered. The final stage shows eventual training data for the relation extractor after SDPs are expanded using self-training.}

\section{Experimental Results}
\label{sec:results} 
% In this Section we evaluate our training strategy. We first describe 
\subsection{Testing Methodology}
We evaluate the relation extractors in the context of an existing Knowledge Base Population system ~\cite{UGent_TAC_2014,UGent_TAC_2015} using the NIST TAC KBP English Slot Filling (ESF) Evaluation from 2012 to 2014. We choose for this evaluation because of the diversity and difficulty of entities in the queries.
%, assuring that a diverse set of test instances is assessed.
% \thomas{rest vd zin heb ik outcommented.}
%\par
In the end-to-end ESF framework, the input to the system is a given entity (the `query'), a set of relations, and a set of articles. The output is a set of slot fills, where each slot fill is a 
%candidate triple in the KB, the first element of which is the query entity. 
triple consisting of two entities (including the query entity) and a relation (from among the given relations) predicted to hold among these entities.

\subsection{Knowledge Base Population System}
Systems participating in the TAC KBP ESF need to handle each task of filling missing slots in a KB. Participants are only provided with one surface-text occurrence of each query entity in a large collection of text provided by the organizers. This means that an information retrieval component is needed to provide the relation extractor with sentences containing candidate answers. Our system performs query expansion using Freebase aliases and Wikipedia pages. Each document containing one of the aliases is parsed and named entities are automatically detected. Persons, organizations, and locations are recognized, and locations are further categorized as cities, states, or countries. Non-entity fillers like titles or charges are tagged using lists and table-lookups.
For further details of the KBP system we refer to \cite{UGent_TAC_2014,UGent_TAC_2015}.

\subsection{Methodologies for Supervision}
\label{sec:annotation_meths}

\revision{
In this section we detail the different procedures for human supervision. Supervision is obtained in two forms: by labeling Shortest Dependency Paths (SDPs) and by labeling single training instances indicated as positive by distant supervision, as either true positives or as false positives (noise). After a background corpus is linked with a knowledge base, phrases containing facts are stored in a database for further feature extraction, post processing, and calculation of feature confidence values.}

\revision{
Our annotators for the labeling of single training instances were undergraduate students from different backgrounds with little or no experience in Machine Learning or Natural Language Processing. First, they were briefed on the semantics of the relation to be extracted using the official TAC KBP guidelines. They were then presented with training instances, \ie phrases from the database.
%The annotators then ran an application which extracts and presents phrases from the database, \ie training instances.
Each instance was shown with entity and subject highlighted and colored. The average time needed to annotate a batch of 2,000 instances was three hours, corresponding to about 5 seconds per instance, including the time needed to read and judge the sentence. As this procedure was relatively expensive (annotators were paid \$15 per hour), only the 15 most frequent relations, strongly influencing the optimal micro-\Fone score shown in Table~\ref{tab:ratio}, were selected. Other relations received between 200 and 1,000 annotations each. 
} 

\revision{In contrast, the time for annotation of the SDPs was limited to merely 5 minutes per relation, during which, on average, 200 SDPs were judged. SDPs were presented in a spreadsheet as a list, and true positives were labeled using a simple checkbox. All SDP annotations were done by a single expert annotator. 
To measure the degree of expertise needed for these annotations, we also assigned a novice annotator (student) with the same task. We measured annotator agreement and time needed for a selection of the relations. For this experiment the student was explained the meaning of dependency paths and the aim of choosing valid SDPs. Several lists of SDPs that the expert was able to label in 5 minutes were presented to the student. For the first two relations the student needed more than 10 minutes to label, but for the subsequent relations, annotation time dropped to 5 minutes per relation, equivalent to the time needed by an expert annotator. 
We measured inter annotator agreement using Cohen's kappa coefficient $\kappa$. Inter-annotator agreement between student and expert was initially moderate ($\kappa = 0.65$) and increased after the student completed lists of SDPs for two relations ($\kappa$ varies between 0.85 and 0.95), indicating a very good agreement. 
% \delete{Coefficients start of at $0.65$ and increase to values between $0.85$ and $0.95$ for subsequent relations.}  
}
%\todo[inline]{@Lucas: The last part, on ``a single (expert) annotator'', is slightly confusing, since you started off by saying that ``annotators are undergraduate students [\ldots] with little or no experience \ldots''. Also, maybe best to include an explicit forward pointer ``(see Section xx)'' to where you provide the agreement results?}

%\subsection{Sampling of Distantly Labeled Instances}
\subsection{Pattern-based Restriction vs. Similarity-based Extension}
As Table~\ref{tab:ratio} shows, applying the manually annotated features as described in Section~\ref{sec:hcoc} often leads to a drastic reduction of training instances, compared to the original weakly labeled training set.
Using similarity metrics described in Section~\ref{sec:semantic}, we again add weakly supervised training data to the filtered data. 
An important question is therefore \revision{how to optimally combine initial reduction with subsequent expanding of the training instances}. Intuitively, one would expect a high-precision-low-recall effect in the extreme case of adding no similar patterns, and a low-precision-high-recall effect when adding all weakly labeled patterns, both leading to a sub-optimal $F_1$ measure. On the other hand, adding a limited amount of similar patterns may increase recall without harming precision too much.  
In this section, we investigate for a selection of relations, how the quality of the training set depends on the  fraction of top similar patterns to extend it with.

In our experimental setup, we start from the training set that only contains the $N_{\textit{filtered}}$ instances that match the manually labeled patterns, gradually adding weakly labeled data, and each time training binary classifiers on the corresponding training set. We chose to let the additional data grow exponentially, which allows studying the effect of adding few extra instances initially, but extending towards the full weakly supervised training set of size $N_{\textit{DS}}$ in a limited number of cases.
More specifically, in $K$ experiments of adding additional instances, the intermediate training set size $N_k$ at step $k$ is given by

\begin{equation}
N_k = N_\textit{filtered} . \left(\frac{N_\textit{DS}}{N_\textit{filtered}}\right)^{k/K} \label{eq:Nk}
\end{equation}

Figure~\ref{fig:samplestrategy} illustrates how an initial training set containing only 5\% of the amount of instances from the full weakly labeled training set, is increased in $K=10$ consecutive experiments.

\begin{figure}
%   \begin{minipage}[c]{0.4\textwidth}
  \centering

% $$ \alpha = exp(\frac{1}{K} ln \frac{N_{DS}}{N_{filtered}})=\sqrt[K]{\frac{N_{DS}}{N_{filtered}}} $$

%   \end{minipage}
%   \hfill
%   \begin{minipage}[c]{0.45\textwidth}
  \include{figs/samples}
	\caption{Example of the proposed sampling strategy for training set sizes, with $N_{filtered}=0.05 N_{DS}$, and in $K=10$ steps.} 
	\label{fig:samplestrategy}
%   \end{minipage}
\end{figure}

\begin{figure}[htbp]
\centering
        \includegraphics[scale=0.27]{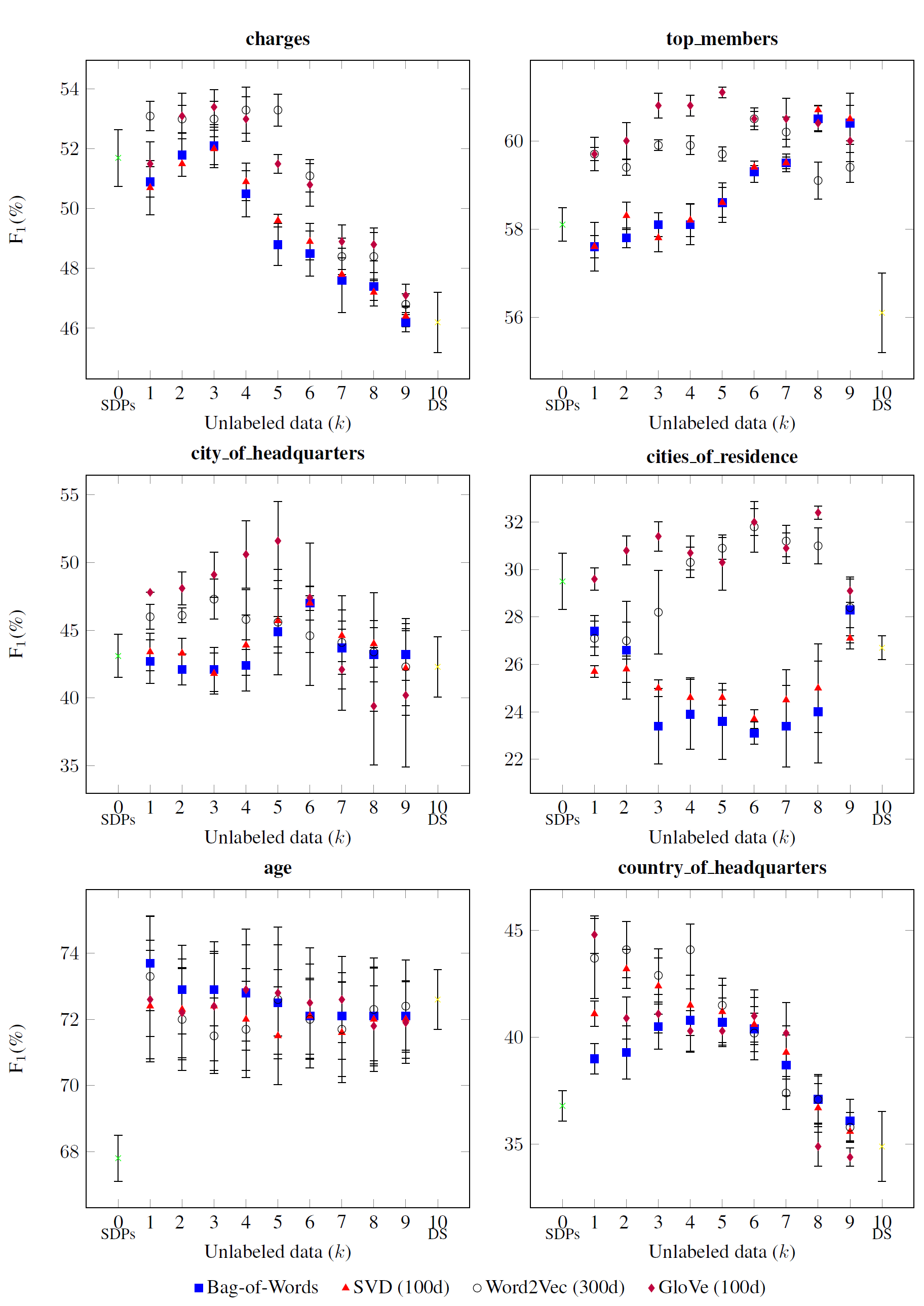}
	\caption{Illustration of the behavior of Semantic Label Propagation for different dimensionality reduction techniques, and different amounts of added weakly labeled data, quantified by $k$ (as in eq.~\ref{eq:Nk}), with $K=10$. $k=0$ corresponds to only accepting manually filtered SDPs, and $k=10$ corresponds to using all weakly labeled (DS) data for training.}
	\label{fig:result}
\end{figure}

% \thomas{(hieronder) dit toch nog wat herschreven om 't meer samenhangend te krijgen.}
Apart from studying the addition of varying amounts of similar patterns, in this section we also investigate the influence of the type of similarity measure used. In Section~\ref{sec:hcoc} we suggested the use of word embeddings, but is there a difference between different types of embeddings? Would embeddings work better than traditional dimension reduction techniques? And would such techniques indeed perform better than the original one-hot vector representations? These questions can be answered by considering several similarity measures.
As a classical baseline, we represent SDPs using the average one-hot or Bag-Of-Words (BOW) representations of the words contained in the SDPs. We also transform the set of one-hot representations using Singular Value Decomposition (SVD)~\cite{deerwester1990indexing} fitted on the complete training set. For representations using the summed average of word embeddings described in Section~\ref{sec:semantic}, we use two sets of pre-trained \textit{Word2Vec} embeddings\footnote[1]{\url{https://code.google.com/p/word2vec/}} (trained on news text) and \textit{GloVe} embeddings\footnote[2]{\url{http://nlp.stanford.edu/projects/glove/}} (trained on Wikipedia text).

%In this Section we study the effect on extraction performance of our labeling strategy and the use of word embeddings compared to one-hot word representations and other dimension reduction techniques from the field of Information Retrieval.
%
% \thomas{uitleg over development set etc, zou ik maar tonen na introductie vd figuur; dan is 't mss duidelijker dat we in de volgende stap (+sectie) hieruit de beste $k$ kiezen om echt op de test set te gaan meten.}
%For different amounts of added weakly supervised data, we measure the optimal \Fone-value of classification on a development set (which consists of training data from 2012 and 2013), in the next Section we evaluate on a held-out test set (which consists of queries from the 2014 TAC ESF task). 
%

%\todo[inline]{@Lucas: Minor remark re formatting of fig.\ \ref{fig:result}: Is it possible to slightly shift the various series, to avoid overlapping error bars? (E.g., in R you can do this with ggplot2 using ``position\_dodge''\ldots) \red{Sorry this wasn't easy using the latex library I used for plots}}

%announce + describe results
\par Figure~\ref{fig:result} shows the effect of adding different amounts of weakly labeled data, for different values of $k$ as in eq.~\ref{eq:Nk} (with $K=10$ steps)  and for similarity measures based on the different types of representations described above. Six frequently occurring relations were selected such that they give an idea of the various forms of behavior that we observed during our investigation of all extracted relations.
%show F1 on development set
The chosen effectiveness measure is the optimal \Fone value of classification on a development set, consisting of training data from 2012 and 2013. \revision{(}In the next Section we \revision{will} evaluate on a held-out test set, which consists of queries from the 2014 TAC ESF task, whereby the optimal value of $k$  and type of dimension reduction is selected based on the development set.\revision{)} Also shown are standard deviations on these optimal \Fone-values, obtained by resampling different positive and negative instances for training the classifier.

Several insights can be gained from Fig.~\ref{fig:result}\revision{:}
\begin{itemize}[nolistsep,leftmargin=*]
\item \revision{\textit{SDPs vs full DS training set:}} We observe that the effect of expanding the initial training set is strongly dependent on the specific relation and the quality of the initial training data. In many cases training data filtered using only highly confident SDPs ($k=0$) generates a better relation extractor than pure DS ($k=K$). This holds for all shown relations, except for the \textit{age} relation. We have to be aware that wrongly annotating an important pattern, or by chance missing any in the top most confident ones, can strongly reduce recall when only using the accepted SDPs. Adding even a small amount of similar patterns may hence result in a steep increase in effectiveness, such as for $k=1$ in the \textit{age} and \textit{country\_of\_headquarters} relations.
\item \revision{\textit{Effect of semantic label propagation:}} When relaxing the filtering (\revision{i.e.,} increasing $k$) by adding unlabeled data, the optimal \Fone  tends to increase until a certain point, and then again drops towards the behavior of a fully DS training set, because the quality or similarity of the added training data declines and too many false positives are re-introduced. The threshold on the amount of added DS instances is thus an important parameter to tune on a development set. For some of the relations there is an optimal amount of added unlabeled data, whereas other relations show no clear optimum and fluctuate between distant and filtered classifiers' values. 
\item \revision{\textit{Impact of dimensionality reduction:}} The use of word embeddings often leads to an improved maximum \Fone value with respect to the BOW-representations or SVD-based dimension reduction. This is for example very clear for the \textit{charges},  \textit{city\_of\_headquarters}, or \textit{cities\_of\_residence} relations, with a slight preference of the \textit{GloVe} embeddings with respect to \textit{Word2Vec} for this application. However, we also noticed that word embeddings are not always better than the BOW or SVD based representations. For example, the highest optimal \Fone for the \textit{age} relation is reached with the BOW model. 
\end{itemize}

%In the next Section we use thresholds on $k$ tuned on the held-out test set, and compare our strategy to different relation extraction models in an End-to-End KBP setting. 
% \thomas{voor mij mag deze laatste zin weg; reeds aangekondigd.}

\subsection{End-to-End Knowledge Base Population Results}
This section presents the results of training binary relation classifiers according to our new strategy for each of the 41 relations of the TAC KBP schema. We tuned hyperparameters on data of the 2012 and 2013 tracks and now test on the last edition of the ESF track of 2014. 

\par Next to the thresholds of choosing the amount of unlabeled data added as discussed previously (i.e., the value of $k$), other parameters include regularization and the ratio between positive and negative instances, which appeared to be an important parameter influencing the confidence of an optimal \Fone value greatly. \revision{Different ratios of negative to positive instances resulted in shifting the optimal trade-off between precision and recall. The amount of available negative training data was on many occasions larger than the available positive.
More negative than positive training data overall appeared to result in lower positive classification probabilities assigned by the classifier to test instances.
Negative instances had to be down-weighted multiple times to prevent the classifier from being too strict and rarely classify a relation as true. For each relation, this parameter was tuned for optimal \Fone value at the $0.5$ probability threshold of the logistic regression classifier. } 
\par We use the official TAC KBP evaluation script which calculates the micro-average of all classifications. All methods are evaluated while ignoring provenances (the character offsets in the documents which contain the justification for extraction of the relation), so as not to penalize any system for finding a new provenance not validated in the official evaluation key. A listing of precision, recall and \Fone for the top 20 most frequently occurring relations in the test set is shown in Table~\ref{tab:resultstable}. 

\par Next to traditional distant supervision (also known as \textit{Mintz++}\cite{mintz2009distant}, indicated as `Distant Supervision' in Table~\ref{tab:resultstable}), we compare our new semi-supervised approach (`Semantic Label Propagation') to a fully supervised classifier trained by manually labeling $50,000$ instances (`Fully Supervised'), and to the classifiers obtained by purely filtering on manually labeled patterns (`SDP Filtered'). We also use the fully supervised classifiers in a traditional self-training scheme, classifying distantly supervised instances in the complete feature space and adding confident instances to the training set (`Self-Training (Instances)'). The supervision needed for these classifiers required far more annotation effort than the feature certainty sampling of Semantic Label Propagation. 

\par
\begin{figure}[H]
\centering
        \includegraphics[scale=0.35]{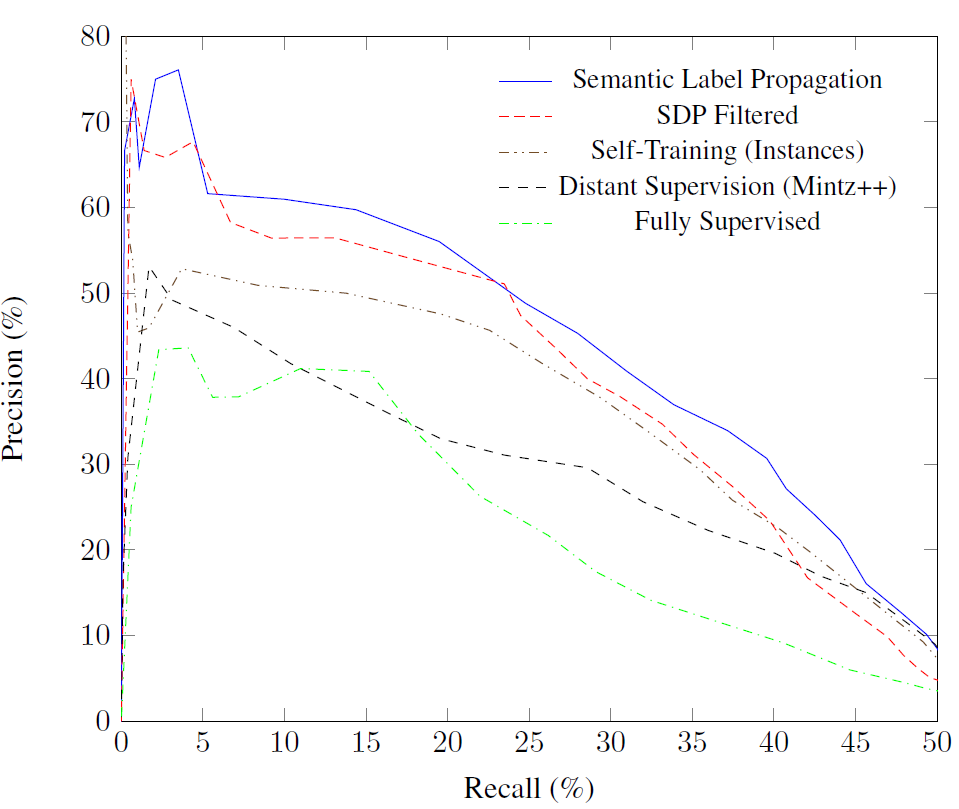}
	\caption{Precision-Recall Graph displaying the output of the TAC KBP evaluation script on different systems, for varying classifier decision thresholds.}
	\label{fig:precrec}
\end{figure}

 The official \Fone value of 36.4\% attained using Semantic Label Propagation is equivalent  to the second best entry out of eighteen submissions to the 2014 ESF track~\cite{surdeanu2014overview}. A relation extractor is but a part of a KBP system and is influenced by each of the other modules (\eg recognition and disambiguation of named entities), which makes it hard to compare to other systems. 
This is the case for the absolute values of Table~\ref{tab:resultstable}, but still, it demonstrates the overall quality of our relation extractors. Especially, our system relying on very limited annotations has a competitive place among systems that rely on many hours of manual feature engineering~\cite{angeli2014stanford}. 
Comparing the results for Semantic Label Propagation with the other approaches shows that the proposed method that combines a small labeling effort based on feature certainty with the Semantic Label Propagation technique, outperforms the DS method, semi-supervision using instance labeling, and full supervision methods. This is also confirmed in Fig.~\ref{fig:precrec}, which shows the trade-off between the precision and recall averaged over all TAC KBP relations for the different methods described above, using the TAC KBP evaluation script (varying the thresholds on classification).
% \todo[inline]{@Lucas: Fig.\ \ref{fig:precrec}: if possible, list legend in order corresponding to positions of the actual curves, i.e., put ``fully supervised'' at the bottom. Also, maybe use color (for the online version)?}
 \begin{sidewaystable}[htbp]
 \caption{Results for Frequent Relations and official TAC-scorer}
 	\centering
   \label{tab:resultstable}
   \include{tables/results}
 \end{sidewaystable}

\revision{One would expect the SDP filtered and fully supervised extractors to attain high precision, but this is not the case for some of the relations. For example, for relation \textit{countries\_of\_residence} recall of these extractors is higher than recall of the SLP method. However, only those precision and recall scores are shown that correspond to the
maximum values for \Fone and while precision could have been higher for these extractors at the cost of lower recall, recall is equally important for this type of evaluation. The SDP filtered and fully supervised extractors are likely to attain high precision values, but this will not compensate for the loss in recall when evaluating \Fone scores.
We conclude by noting that the results may also be influenced to peculiarities of the data. Entities chosen by TAC may not always be representative for the majority of persons or organizations in the training data: TAC entities are in many cases more difficult than the average entity from the training set and the most common way of expressing a relationship for these entities might not be present in the test set.}

% \tikzsetnextfilename{precrec}

%Figure~\ref{fig:precrec} shows the trade-off between precision and recall for the different methods described above, using the TAC KBP evaluation script, by varying the thresholds on classification.
%This confirms that feature certainty labeling with semi-supervision, \ie the Semantic Label Propagation method introduced in this paper, consistently outperforms the other methods DS, semi-supervision using instance labeling, and full supervision.
% Overall performance Mintz++ is especially hurt by scoring bad on one of the relations most present in the test set. 
\revision{
\subsection{2015 TAC KBP Cold Start Slot Filling} 
The Slot filling task in TAC KBP in 2015 was organized as part of the Cold Start Slot Filling track, where the goal is to search the same document collection to fill in values for specific slots for specific entities, and in a second stage fill slots for answers of the first stage. In the authors' TAC KBP 2015 submission~\cite{UGent_TAC_2015}, some of the ideas presented in this paper were applied, leading to a second place in the Slot Filling Variant. 
The results showed the influence of a clean training set and the effectiveness of self-training. A top-performing entry was based on the DeepDive-like database system~\cite{zhang2015deepdive} and training set filtering. We note that the idea of self-training using a first stage high-precision classifier was also included in this DeepDive system, independently of the work presented in this paper. Neural Architectures for relation extractors were also included in ensembles of some systems and found to be important extractors but a selection of our Linear classifiers in combination with a careful filtering of distantly supervised training data was shown to outperform ensembles of Linear and Neural classifiers.
}

%\section{Conclusions and Future Work}\label{sec:conclusions}
\section{Conclusions}\label{sec:conclusions}
%We presented a strategy to upgrade distantly supervised training data for relation extraction. Feature labeling after an initial training removes noise and focuses the classifier but ignores the diversity of the training set generated by distant supervision. We relax the assumption of filtering training data for highly confident Shortest Dependency Paths by measuring similarities between these and the remaining distantly supervised training data. We found that a dimension reduction by a simple linear combination of the word embeddings contributing in the relation pattern are effective representations to propagate labels from the supervised to the weakly supervised instances. Tuning a threshold parameter for similarity creates an improved training set for relation extraction.
%
The overall aim of our proposed strategy in building relation extractors in a closed IE setting (i.e., to extract a priori specified relations) is to maximize their performance using minimal (human) annotation effort. Our key ideas to achieve that are:
\inlinelist{(\roman*)}{
\item \label{it:idea-ds} \textit{distant supervision (DS):} use known relation instances from a knowledge base to automatically generate training data,
\item \label{it:idea-fa} \textit{feature annotation:} rather than labeling instances, annotate features (e.g., text patterns expressing a relationship), selected by means of an active learning criterion, and \item \label{it:idea-semantic} \textit{semantic feature space representation:} use compact semantic vector spaces to detect additional, semantically related patterns that do not occur in the thus far selected training data, leaving useful patterns undetected otherwise.
}

Thus, we address the problem of noisy training data obtained through DS by expanding the key idea of automatically filtering of the training data to increase precision (see \cite{sterckx2014using}). Specifically, to improve recall, we introduce the semi-supervised Semantic Label Propagation method, that allows to relax the pattern-based filtering of the DS training data by again including weakly supervised items that are sufficiently ``similar'' to highly confident instances.
We found that a simple linear combination of the embeddings of the words contributing to the relation pattern is an effective dimension reduction technique to obtain representations for propagating labels from supervised to weakly supervised instances. Tuning a threshold parameter for similarity creates an improved training set for relation extraction. 

The main contributions of this paper to the domain of closed relation extraction, are 
\inlinelist{(\roman*)}{
\item the novel methodology of filtering an initial DS training set, where we motivated and demonstrated the effectiveness of an almost negligible manual annotation effort, and 
\item the Semantic Label Propagation model for again expanding the filtered set in order to increase diversity in the training data.} 

%Note that \chris{Say something about what ideas in \ref{it:idea-ds}--\ref{it:idea-al} are new, or that their combination is novel (and/or the quantitative identification of how helpful they are)?}\ldots

\par We evaluated our classifiers on the knowledge base population task of TAC KBP and show the competitiveness with respect to established methods that rely on a much heavier annotation cost.

\section*{References}
\bibliography{mybibfile}

\end{document}

%% file: figs/ds_example.tex
\tikzset{
    state/.style={
           rectangle,
           draw=black, very thick,
           minimum height=2em
           },
}

% For scaling: thanks to http://tex.stackexchange.com/questions/26846/how-to-scale-a-tikzpicture-including-texts
\begin{tikzpicture}[->,>=stealth',scale=0.85, every node/.style={transform shape}]

 % First node
 % Use previously defined 'state' as layout (see above)
 % use tabular for content to get columns/rows
 % parbox to limit width of the listing
 \node[cylinder,draw=black,thick,aspect=0.1,
        minimum height=2cm,minimum width=6.6cm,
        shape border rotate=90] (KB) 
 {
 
  \parbox{6.7cm}{
  \centering
 \textbf{Knowledge Base}
  \begin{tabular}{l|l|l}
   Relation ($r$) & Entity 1 ($e_1$) &    \small Entity 2 ($e_2$) \\\hline
   $born\_in$ & Barrack Obama & U.S. \\
   $spouse$ & Barrack Obama & Michelle \\
   \ldots & \ldots & \ldots
  \end{tabular}
  }
};

 \node[state,       % layout (defined above)
 node distance=9.0cm,     % distance to QUERY
 text width=10cm,        % max text width
 right of=KB,        % Position is to the right of QUERY
 yshift=+0cm] (DOCS)    % move 3cm in y
 {%   
   
   \footnotesize
 \begin{tabular}{ll}     % content
 \textbf{Mentions in free text}& \textbf{True +?}\\
 \hline
\textbf{Barack} was born in Honolulu, Hawaii, \textbf{U.S.} & \ding{51}\\
\textbf{Barrack Obama }ended \textbf{U.S.} military involvement in the Iraq War.& \ding{55} \\
\textbf{Michelle} and \textbf{Barack} are visiting Cuba.& \ding{55}\\
Barack and his wife \textbf{Michelle} are meeting with Xi Jinpeng & \ding{51}  

 \end{tabular}
 };

 % draw the paths and and print some Text below/above the graph

\path  (KB) edge                    (DOCS)

 ;
\end{tikzpicture}

%% file: figs/cumprob.tex
\begin{tikzpicture}
	\begin{axis}[
		enlarge x limits=false,
        ymode=log,
        xmode=log,
        ylabel=  Frequency in training set ,
		xlabel=Normalized sorted rank,
        xmax =1,
        xmin=0,
        ymin=0,
        ymax=1,
        legend pos=north east,
        legend style={ draw=none,font=\small},  
        mark size=1.5pt,
        width=6.5cm,
        every axis plot post/.append style={mark=none},
        cycle list name=linestyles]
	
    \addplot
    table[x=x,y=y]
	{data/freq_org_top_members_employees.dat};

	\addplot
    table[x=x,y=y]
	{data/freq_per_date_of_death.dat};

	\addplot
    table[x=x,y=y]
	{data/freq_per_employee_or_member_of.dat};

	\legend{top\_member,date\_of\_death,employee\_of}
	\end{axis}    
\end{tikzpicture}%

%% file: figs/conf.tex
\begin{tikzpicture}
	\begin{axis}[
		enlarge x limits=false,
        ylabel=  Confidence,
		xlabel= Sorted rank,
        legend pos=north east,
        legend style={ draw=none,font=\small},  
        mark size=1.2pt,
        cycle list name=mark list,
        width=6.5cm
        %every axis plot post/.append style={mark=none},
        ]
	\addplot[only marks,mark=*]
    table[x=x,y=y]
	{data/contrue_schools.dat};
     
	\addplot[only marks,mark=square*]
    table[x=x,y=y]
	{data/contrue_topmembers.dat};
        
	\addplot[only marks,mark=triangle*]
    table[x=x,y=y]
	{data/contrue_founded_by.dat};

   	\addplot[no markers]
    table[x=x,y=y]
	{data/con_schools.dat};

	\addplot[no markers]
    table[x=x,y=y]
	{data/con_topmembers.dat};
	\legend{True Positive Patterns}
   
   	\addplot[no markers]
    table[x=x,y=y]
	{data/con_founded_by.dat};
	\legend{True Positive Patterns}

   \legend{schools\_attended,top\_member,founded\_by}

	\end{axis}
  
\end{tikzpicture}

%% file: figs/samples.tex
	\begin{tikzpicture}[scale=0.65,baseline]
	\begin{axis}[xlabel=k,ylabel=Fraction of DS Trainingset Included (\%),xmin=0,xmax=10,ymin=0,ymax=100,xticklabels={0\\SDP, 1 , 2 ,3,4,5,6,7,8,9, 10\\DS},xtick={0,...,10}, x tick label style={rotate=0,anchor=north,align=center},only marks]
			\addplot
            table[x=x,y=y ]{
x     y      
0	5
1	6.7464142384
2	9.1028210151
3	12.2822802612
4	16.5722700867
5	22.360679775
6	30.1708816827
7	40.7090531537
8	54.9280271653
9	74.1134449107
10	100
	};
		\end{axis}
	\end{tikzpicture}
    

%% file: tables/results.tex
\footnotesize
\begin{tabular}{l@{\hskip 3 \tabcolsep}
*{3}{c}@{\hskip 3 \tabcolsep}
*{3}{c}@{\hskip 3 \tabcolsep}
*{3}{c}@{\hskip 3 \tabcolsep}
*{3}{c}@{\hskip 3 \tabcolsep}
*{3}{c}}
% lcccc@{\hskip 3 \tabcolsep}ccc@{\hskip 3 \tabcolsep}ccc@{\hskip 3 \tabcolsep}ccc}
\toprule  
   & \multicolumn{3}{@{}c@{\hskip 3 \tabcolsep}}{\parbox{2.5cm}{\centering \textbf{Distant Supervision}\\\textbf{(Mintz++)}}}
   & \multicolumn{3}{@{}c@{\hskip 3 \tabcolsep}}{\textbf{SDP Filtered}}
   & \multicolumn{3}{@{}c@{\hskip 3 \tabcolsep}}{\textbf{Fully Supervised}}
   & \multicolumn{3}{@{}c@{\hskip 3 \tabcolsep}}{\parbox{2.5cm}{\centering \textbf{Self-Training}\\ \textbf{(Instances)}}}
   & \multicolumn{3}{@{}c}{\parbox{2.5cm}{\centering \textbf{Semantic}\\ \textbf{Label Propagation}}} \\
   \cmidrule(lr{\dimexpr 3\tabcolsep+0.5em}){2-4}
   \cmidrule(lr{\dimexpr 3\tabcolsep+0.5em}){5-7}
   \cmidrule(lr{\dimexpr 3\tabcolsep+0.5em}){8-10}
   \cmidrule(lr{\dimexpr 3\tabcolsep+0.5em}){11-13}
   \cmidrule(l{0.2em}r{1em}){14-16}
   %\midrule
 Relation & P &   R & \Fone & P &   R & \Fone & P &   R & \Fone & P &   R & \Fone  & P &   R & \Fone \\ 
    \midrule
\textbf{title}				&	22.3	&	58.8	&	32.3	&	36.1	&	39.1	&	37.5	&	28.0	&	61.1	&	38.4	&	36.5	&	43.2	&	\textbf{39.6}	&	37.3	&	41.2	&	\revision{39.2}	\\
\textbf{top\_members\_employees}				&	50.6	&	63.4	&	56.3	&	51.3	&	63.4	&	56.7	&	62.6	&	53.9	&	57.9	&	56.3	&	63.4	&	59.6	&	63.5	&	62.5	&	\textbf{63.0}	\\
\textbf{employee\_or\_member\_of}				&	31.4	&	34.0	&	32.6	&	33.8	&	51.0	&	\textbf{40.7}	&	23.5	&	45.7	&	31.0	&	32.2	&	40.4	&	35.8	&	27.9	&	51.0	&	\revision{36.1}	\\
\textbf{age}				&	71.6	&	72.5	&	72.0	&	75.6	&	70.0	&	72.7	&	68.0	&	62.5	&	64.9	&	73.6	&	70.0	&	71.8	&	68.8	&	82.5	&	\textbf{75.0}	\\
\textbf{origin}				&	100.0	&	23.0	&	37.4	&	28.5	&	80.0	&	42.0	&	29.4	&	66.6	&	40.8	&	27.5	&	73.3	&	40.0	&	31.7	&	86.6	&	\textbf{46.4}	\\
\textbf{countries\_of\_residence}				&	100.0	&	23.0	&	37.4	&	22.4	&	84.6	&	35.4	&	22.2&	92.3&	35.8 &	50.0	&	38.4	&	\textbf{43.4}	&	35.2	&	46.1	&	39.9	\\
\textbf{charges}				&	45.0	&	52.9	&	48.6	&	40.9	&	52.9	&	46.1	&	70.4	&	44.1	&	\textbf{54.2}	&	47.6	&	58.8	&	52.6	&	44.3	&	68.1	&	53.7	\\
\textbf{cities\_of\_residence}				&	22.9	&	45.8	&	30.5	&	31.5	&	25.0	&	27.9	&	11.2	&	62.5	&	19.0	&	36.3	&	16.6	&	22.8	&	34.4	&	41.6	&	\textbf{37.7}	\\
\textbf{cause\_of\_death}				&	30.7	&	36.3	&	33.3	&	29.4	&	45.4	&	35.7	&	28.3	&	31.8	&	29.9	&	37.5	&	27.2	&	31.5	&	33.3	&	45.4	&	\textbf{38.4}	\\
\textbf{spouse}				&	50.0	&	45.4	&	47.6	&	50.0	&	45.4	&	47.6	&	75.0	&	27.2	&	39.9	&	35.7	&	45.4	&	40.0	&	71.4	&	45.4	&	\textbf{55.5}	\\
\textbf{city\_of\_death}				&	100.0	&	16.6	&	28.5	&	14.2	&	16.6	&	15.3	&	5.2	&	100.0	&	9.9	&	20.0	&	16.6	&	18.1	&	20.0	&	33.3	&	\textbf{25.0}	\\
\textbf{country\_of\_headquarters}				&	22.7	&	41.6	&	29.4	&	62.5	&	41.6	&	\textbf{50.0}	&	25.0	&	50.0	&	33.3	&	100.0	&	25.0	&	40.0	&	100.0	&	33.3	&	\textbf{50.0}	\\
\textbf{date\_of\_death}				&	66.6	&	50.0	&	\textbf{57.1}	&	66.6	&	50.0	&	\textbf{57.1}	&	50.0	&	25.0	&	33.3	&	66.6	&	50.0	&	\textbf{57.1}	&	66.6	&	50.0	&	\textbf{57.1}	\\
\textbf{(per:)parents}				&	37.0	&	50.0	&	42.5	&	42.1	&	40.0	&	41.0	&	37.5	&	15.0	&	21.4	&	34.6	&	45.0	&	39.1	&	40.9	&	45.0	&	\textbf{42.9}	\\
\textbf{(org:)alternate\_names}				&	20.0	&	28.5	&	23.5	&	18.7	&	85.7	&	30.7	&	20.0	&	28.5	&	23.5	&	16.2	&	85.7	&	27.2	&	19.3	&	85.7	&	\textbf{31.5}	\\
\textbf{statesorprovinces\_of\_residence}				&	50.0	&	55.5	&	\textbf{52.6}	&	50.0	&	44.4	&	47.0	&	53.5	&	44.4	&	48.5	&	45.4	&	55.5	&	49.9	&	50.0	&	44.4	&	47.0	\\
\textbf{founded\_by}				&	53.8	&	43.7	&	48.2	&	80.0	&	50.0	&	61.5	&	75.0	&	37.5	&	50.0	&	62.5	&	62.5	&	62.5	&	81.8	&	56.2	&	\textbf{66.6}	\\
\textbf{children}				&	21.4	&	27.2	&	24.0	&	35.7	&	45.4	&	40.0	&	50.0	&	9.2	&	15.5	&	27.2	&	27.2	&	27.2	&	38.4	&	45.4	&	\textbf{41.6}	\\
\textbf{city\_of\_headquarters}				&	42.8	&	100.0	&	59.9	&	46.1	&	66.6	&	54.5	&	36.3	&	88.8	&	51.5	&	46.6	&	77.7	&	58.3	&	71.4	&	55.5	&	\textbf{62.5}	\\
\textbf{siblings}				&	100.0	&	28.5	&	\textbf{44.4}	&	100.0	&	28.5	&	\textbf{44.4}	&	100.0	&	14.2	&	24.9	&	66.6	&	28.5	&	39.9	&	100.0	&	28.5	&	\textbf{44.4}	\\
\textbf{(org:)parents}				&	33.3	&	33.3	&	33.3	&	33.3	&	66.6	&	\textbf{44.4}	&	33.3	&	33.3	&	33.3	&	33.3	&	33.3	&	33.3	&	33.3	&66.6&	\revision{\textbf{44.4}}
	\\
\midrule
\textbf{Official TAC Scorer} \textbf{( Micro-\Fone)} & 29.3&   28.1&   28.7& 35.5& 33.7&   34.7     &  22.7&   26.0&   24.3  &  37.5 &  29.4  &  33.0&  36.9 & 35.9 & \textbf{36.4}
\\ 
\bottomrule
% 0.3305227655986509 0.37789203084832906 0.2937062937062937
%0.3638 0.3680	0.359
\end{tabular}
\normalsize